\documentclass[10pt,twocolumn,letterpaper]{article}

\usepackage{ijcb}
\usepackage{times}
\usepackage{epsfig}
\usepackage{graphicx}
\usepackage{multirow}
\usepackage{amsmath}
\usepackage{amssymb}
\usepackage{color,soul}
\usepackage{hyperref}
\hypersetup{
  colorlinks, linkcolor=red
}

\ijcbfinalcopy 

\ifijcbfinal\fi
\begin{document}

\title{RidgeBase: A Cross-Sensor Multi-Finger Contactless Fingerprint Dataset}

\author{Bhavin Jawade, Deen Dayal Mohan, Srirangaraj Setlur,  Nalini Ratha, Venu Govindaraju\\
Computer Science and Engineering \\
University at Buffalo, SUNY\\
{\tt\small \{bhavinja, dmohan, setlur, nratha, govind\}@buffalo.edu}
}

\maketitle
\thispagestyle{empty}

\begin{abstract}
Contactless fingerprint matching using smartphone cameras can alleviate major challenges of traditional fingerprint systems including hygienic acquisition, portability and presentation attacks. However, development of practical and robust contactless fingerprint matching techniques is constrained by the limited availability of large scale real-world datasets. To motivate further advances in contactless fingerprint matching across sensors, we introduce the RidgeBase benchmark dataset. RidgeBase consists of more than 15,000 contactless and contact-based fingerprint image pairs acquired from 88 individuals under different background and lighting conditions using two smartphone cameras and one flatbed contact sensor. Unlike existing datasets, RidgeBase is designed to promote research under different matching scenarios that include Single Finger Matching and Multi-Finger Matching for both contactless-to-contactless (CL2CL) and contact-to-contactless (C2CL) verification and identification. Furthermore, due to the high intra-sample variance in contactless fingerprints belonging to the same finger, we propose a set-based matching protocol inspired by the advances in facial recognition datasets. This protocol is specifically designed for pragmatic contactless fingerprint matching that can account for variances in focus, polarity and finger-angles. We report qualitative and quantitative baseline results for different protocols using a COTS fingerprint matcher (Verifinger) and a Deep CNN based approach on the RidgeBase dataset. The dataset can be downloaded here: \url{https://www.buffalo.edu/cubs/research/datasets/ridgebase-benchmark-dataset.html}
\end{abstract}

\vspace{-1em}

\section{Introduction}

Fingerprints are one of the most widely used biometric modalities. Recent works \cite{ispfdv1, ispfdv2, lin2018matching, Lin2018ContactlessAP}in fingerprint recognition have focused their attention on contactless fingerprint matching owing to various benefits over contact-based methods. Traditional fingerprint sensors which require a physical contact with the acquisition surface elevate the risk of spread of contagious diseases. Furthermore, contact with a fingerprint platen leaves a latent impression which can be captured for fingerprint presentation attacks. Contactless fingerprint matching using smartphone cameras alleviates these concerns while also making the acquisition process easier, faster, and portable. \footnote{Code for the acquisition app can be accessed here: \url{https://github.com/bhavinjawade/FingerprintCameraApp}}

Despite its apparent benefits, performing robust contactless fingerprint matching is more challenging than traditional fingerprint matching. The major challenges with contactless fingerprint matching include: out-of-focus image acquistion, lower contrast between ridges and valleys, variations in finger-angle, and perspective distortion. A resilient contactless fingerprint acquisition system must overcome these challenges while being capable of performing both contactless to contactless (CL2CL) and contact to contact-less (C2CL) fingerprint matching. 

\begin{figure}[t]
\includegraphics[scale=0.35]{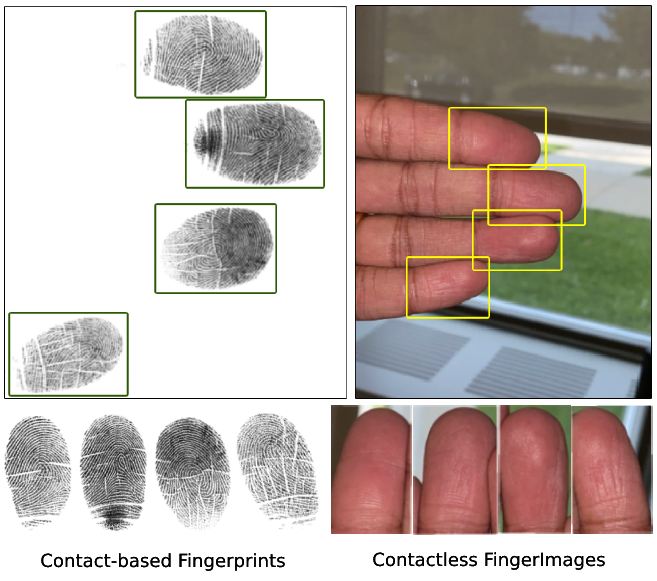}
\caption{Sample contactless and contact-based acquisition images along with segmented fingerprint images from the dataset.}
\end{figure}

\begin{table}[t]
\begin{tabular}{l|l}
\hline
\textbf{Property}                                   & \textbf{Count} \\ \hline
Number of Smartphones                      & 2     \\
Backgrounds and lighting conditions        & 3     \\
Number of unique fingers                   & 704   \\
Number of unique hands (four fingers)      & 176   \\
Number of four-finger contactless images   & 3374  \\
Number of four-finger contact-based images & 280   \\
Number of contactless finger images        & 13484 \\
Number of contact-based fingerprint images & 1120   \\ \hline
\end{tabular}
\vspace{0.5em}
\label{tab:stats}
\caption{Summary of dataset statistics.}
\vspace{-1.0em}
\end{table}

Early attempts \cite{polyudataset} at contact to contactless fingerprint matching proposed datasets that were collected in specialized environmental settings. Other works \cite{ispfdv1, ispfdv2} proposed smartphone captured finger-selfies under different background and lighting conditions. In order to develop robust contactless fingerprint matching that can be used in practically viable systems, research datasets should preferably include:
(i) Images acquired in different lighting conditions and backgrounds.
(ii) Different camera sensors.
(iii) Multi-finger (Four finger) images.
(iv) Images acquired in Unconstrained or semi-constrained settings.
and (vi) Large number of images with high resolution.


Existing contactless fingerprint matching datasets are found to be limited in their scope because they do not meet one or more of the aforementioned conditions. In this paper, we propose RidgeBase,  a large-scale multi-finger contactless and contact-based fingerprint dataset obtained using multiple sensors in diverse environmental conditions and backgrounds. Over 3500 contactless and contact-based \textit{four-finger} images are obtained from 88 subjects in multiple sessions. To enable finger-to-finger matching, the four-finger images are further split into single-finger images. In all (including single finger and four-finger), RidgeBase consists of 17,784 contactless and contact-based images captured in self-operated mode by participants using two smartphone cameras and contact-based sensors. 

Contactless finger images of the same finger distal acquired using a smartphone camera contain higher degree of intra-class variance (due to focus, contrast and angles distortions) when compared to traditional contact-based images. Capturing multiple images of the same finger at acquisition and inference time can improve matching performance. We observe that existing works use traditional sample based evaluation protocol for contactless fingerprint matching. Inspired by the Janus Benchmark Dataset's \cite{januscvpr} evaluation protocol for face recognition, we propose a set-based evaluation protocol for RidgeBase along with other matching protocols. This comprehensive evaluation suite consisting of three tasks:  1. Single finger-to-finger matching 2. Four-finger image matching and 3. Set-based fingerprint matching. Each evaluation task is performed for both contactless (CL2CL) and cross-sensor (C2CL) fingerprint matching, thereby facilitating the development of a robust cross-sensor fingerprint matching framework.

The key contributions of this work are summarized below:
\begin{enumerate}
\item Collected a new cross-sensor fingerprint dataset which overcomes many drawbacks of existing datasets, and is designed to promote practical contactless fingerprint matching research.
\item Proposed a novel fingerprint distal labeling heuristic algorithm to generate pseudo labels for training a Faster-RCNN based object detector for distal segmentation. We also provide fingerprint quality metric (NFIQ) distribution on the RidgeBase dataset.
\item Developed an extensive tasks and protocols suite for RidgeBase that emulates real-world scenarios and ensures reproducibility.
\item Finally, we report baseline results on the RidgeBase dataset using a state-of-the-art commercial-off-the-shelf fingerprint matcher (Verifinger 12.0) and a DeepCNN \cite{wifs_bhavin} based method.
\end{enumerate}

\section{Related Works}

Automatic fingerprint matching is a well researched area. Recently, fingerprint interoperability, especially C2CL matching has gained popularity. In this section we will discuss relevant datasets and prior methods for C2CL matching.

NISTIR 8307 \cite{NISTreport} performed an Interoperability Assessment with data collected from 200 Federal employees to evaluate various existing contactless fingerprint acquisition devices and smartphone apps. They observed that performance of DUTs (Devices under test) can be categorized into three tiers: where the best performing tier consists of contact based devices, the middle tier consists of stationary contactless devices and the worst performing tier consists of smartphone based contactless fingerprint matching apps. Furthermore, NISTIR 8307 \cite{NISTreport} also concluded that multi-finger acquisition of contactless fingerprints increases the performance of contactless matching, thereby enhancing potential operational utility. This further corroborates the importance of our publicly released dataset for research in multi-finger smartphone based contactless fingerprint matching.

Ross et al \cite{ross2004biometric} were among the first to draw attention to problems with biometric sensor interoperability in the context of fingerprints, and they observed that the need for sensor interoperability was paramount as it significantly impacted the usability of a biometric matching system with cross-sensor performance being notably worse. \cite{modi2008analysis} addresses the problem of interoperability by analyzing fingerprint data from  9 different sensors. These methods while promising, only focused on data acquired using contact based fingerprint sensors. Lee et al \cite{lee2006preprocessing} proposed methods to process images captured using mobile phones. \cite{lee2008recognizable} used gradient information coherence to perform finger quality estimation. Recent methods on contactless fingerprint recognition have focused on three main sub areas namely, segmentation of area of interest, enhancement of the segmented area and representation learning for matching. \cite{malhotra2020matching} developed a segmentation model using saliency map and skin color following which Grosz et al. \cite{grosz2021c2cl} proposed a U-Net based autoencoder to avoid failure cases in the presence of complex backgrounds. Compared to contact fingerprint data, contactless fingerprint images suffer from various distortions. Lin et al. \cite{lin2016improving}\cite{lin2018matching} have proposed algorithms that correct non-linear deformations as well as methods for generalized distortion correction based on a robust thin-
spline plate mode. Once the image enhancement is performed, the fingerprint matching is generally done using either minutiae based methods or deep learning based methods. \cite{lin2018cnn} proposed the use of a Siamese CNN architecture for matching contact to contactless fingerprints. Malhotra et al.\cite{malhotra2020matching} designed a network to extract features which preserve the mulit-orientation and multi-scale information of the fingerprint. 
Table \ref{dataset} shows the comparisons of our datasets to other contactless datasets present in the literature. Although Deb et al. 2018 \cite{deb2018} and Wild et al. 2019 \cite{wild2019} have multi-finger images, they do not have the environmental and background variations present in our dataset. Furthermore the number of unique samples (four-finger and single images) are small compared to the proposed dataset.

\begin{table*}[t]
\label{dataset}
\begin{tabular}{l|c|c|c|c|ccl}
\hline
\multicolumn{1}{c|}{\multirow{5}{*}{\textbf{Database}}} & \multirow{5}{*}{\textbf{\begin{tabular}[c]{@{}c@{}}Number \\ of \\ unique \\ fingers\end{tabular}}} & \multirow{5}{*}{\textbf{\begin{tabular}[c]{@{}c@{}}Number \\ of \\ distal \\ images\end{tabular}}} & \multirow{5}{*}{\textbf{\begin{tabular}[c]{@{}c@{}}Multi-\\ finger \\ Images\end{tabular}}} & \multirow{5}{*}{\textbf{\begin{tabular}[c]{@{}c@{}}Full-\\ Finger\\ Images\end{tabular}}} & \multicolumn{3}{c}{\multirow{3}{*}{\textbf{Variations}}}                                                                                                                       \\
\multicolumn{1}{c|}{}                                   &                                                                                                     &                                                                                                    &                                                                                             &                                                                                           & \multicolumn{3}{c}{}                                                                                                                                                           \\
\multicolumn{1}{c|}{}                                   &                                                                                                     &                                                                                                    &                                                                                             &                                                                                           & \multicolumn{3}{c}{}                                                                                                                                                           \\ \cline{6-8} 
\multicolumn{1}{c|}{}                                   &                                                                                                     &                                                                                                    &                                                                                             &                                                                                           & \multicolumn{1}{c|}{\multirow{2}{*}{\textbf{E*}}} & \multicolumn{1}{c|}{\multirow{2}{*}{\textbf{B*}}} & \multicolumn{1}{c}{\multirow{2}{*}{\textbf{Sensors}}} \\
\multicolumn{1}{c|}{}                                   &                                                                                                     &                                                                                                    &                                                                                             &                                                                                           & \multicolumn{1}{c|}{}                                      & \multicolumn{1}{c|}{}                                     & \multicolumn{1}{c}{}                                  \\ \hline
Lee et al., 2005 \cite{lee2006preprocessing}                                       & 168                                                                                                 & 840                                                                                                &                                                                                             &                                                                                           & \multicolumn{1}{c|}{}                                      & \multicolumn{1}{c|}{}                                     & 1.3M pixel CMOS camera                                \\ \hline
Lee et al., 2008 \cite{lee2008recognizable}                                        & 120                                                                                                 & 1200                                                                                               &                                                                                             &                                                                                           & \multicolumn{1}{c|}{}                                      & \multicolumn{1}{c|}{}                                     & Camera                                                \\ \hline
Stein. et al. 2012 \cite{stein2012}                                     & 82                                                                                                  & 492                                                                                                &                                                                                             &                                                                                           & \multicolumn{1}{c|}{}                                      & \multicolumn{1}{c|}{}                                     & Nexus S, Galaxy Nexus                                 \\ \hline
Derawi et al. 2012 \cite{Derawi2012}                                     & 220                                                                                                 & 1320                                                                                               &                                                                                             &                                                                                           & \multicolumn{1}{c|}{}                                      & \multicolumn{1}{c|}{}                                     & Nokia N95, HTC Desire                                 \\ \hline
Li et al., 2012 \cite{li2012testing}                                        & 100                                                                                                 & 2100                                                                                               &                                                                                             &                                                                                           & \multicolumn{1}{c|}{\checkmark}             & \multicolumn{1}{c|}{\checkmark}            & -                                                     \\ \hline
Stein. et al. 2013 \cite{Stein2013VideobasedFR}                                     & 74                                                                                                  & 990                                                                                                &                                                                                             &                                                                                           & \multicolumn{1}{c|}{\checkmark}             & \multicolumn{1}{c|}{\checkmark}            & Galaxy Nexus, Galaxy S3                               \\ \hline
Tiwari et al. 2015 \cite{tiwari2015}                                    & 50                                                                                                  & 156                                                                                                &                                                                                             &                                                                                           & \multicolumn{1}{c|}{}                                      & \multicolumn{1}{c|}{}                                     & Samsung Note GT-N7000                                 \\ \hline
Sankaran et al. 2015 \cite{ispfdv1}                                  & 128                                                                                                 & 5100                                                                                               &                                                                                             &                                                                                           & \multicolumn{1}{c|}{\checkmark}             & \multicolumn{1}{c|}{\checkmark}            & iPhone 5                                              \\ \hline
Lin et al. 2018 \cite{lin2018matching}                                        & 300                                                                                                 & 1800                                                                                               &                                                                                             &                                                                                           & \multicolumn{1}{c|}{}                                      & \multicolumn{1}{c|}{}                                     & camera, URU 4000                                      \\ \hline
Deb et al. 2018 \cite{deb2018}                                       & 1236                                                                                                & 2472                                                                                               & \checkmark                                                                   & \checkmark                                                                 & \multicolumn{1}{c|}{}                                      & \multicolumn{1}{c|}{}                                     & 
\shortstack{SilkID, Guardian 200 \\ Redmi Note 4}
                    \\ \hline
Chopra et al. 2018  \cite{Chopra2018UnconstrainedFD}                                   & 230                                                                                                 & 3450                                                                                               &                                                                                             &                                                                                           & \multicolumn{1}{c|}{\checkmark}             & \multicolumn{1}{c|}{\checkmark}            & Multiple Smartphones                                  \\ \hline
Wasnik et al. 2018   \cite{Wasnik2018BaselineEO}                                  & 48                                                                                                  & 720                                                                                                &                                                                                             &                                                                                           & \multicolumn{1}{c|}{}                                      & \multicolumn{1}{c|}{}                                     & iPhone 6                                              \\ \hline
Lin et.al 2018  \cite{Lin2018ContactlessAP}                                       & 300                                                                                                 & 3920                                                                                               &                                                                                             &                                                                                           & \multicolumn{1}{c|}{}                                      & \multicolumn{1}{c|}{}                                     & -                                                     \\ \hline
Wild et al. 2019  \cite{wild2019}                                      & 108                                                                                                 & 4310                                                                                               & \checkmark                                                                   & \checkmark                                                                 & \multicolumn{1}{c|}{}                                      & \multicolumn{1}{c|}{}                                     & LG Flex 2, Note 4, ARH AFS510                         \\ \hline
Malhotra et al. 2020 \cite{ispfdv2}                                    & 304                                                                                                 & 19456                                                                                              &                                                                                             &                                                                                           & \multicolumn{1}{c|}{\checkmark}             & \multicolumn{1}{c|}{\checkmark}            & 
\shortstack{OnePlus One, MicroMax Knight \\ Secugen}
\\ \hline
\multicolumn{1}{l|}{\textbf{RidgeBase (Ours)}}                                       & \textbf{704}                                                                                        & \textbf{14328}                                                                                     & \textbf{\checkmark}                                                          & \textbf{\checkmark}                                                        & \multicolumn{1}{c|}{\textbf{\checkmark}}    & \multicolumn{1}{c|}{\textbf{\checkmark}}   & \textbf{\shortstack{Google Pixel 5, iPhone 11 \\ Futronic FS64}}     \\ \hline
\end{tabular}
\vspace{0.5em}
\caption{A quantitative comparison of contactless fingerprint datasets in literature. E*: Variations in Environment (Lighting conditions), B*: Variations in Background. A similar compilation can be seen in \cite{ispfdv2}}.
\end{table*}

\section{RidgeBase Dataset}

\subsection{Collection Methodology}

\begin{figure}
\label{appscreenshots} 
\includegraphics[scale=0.33]{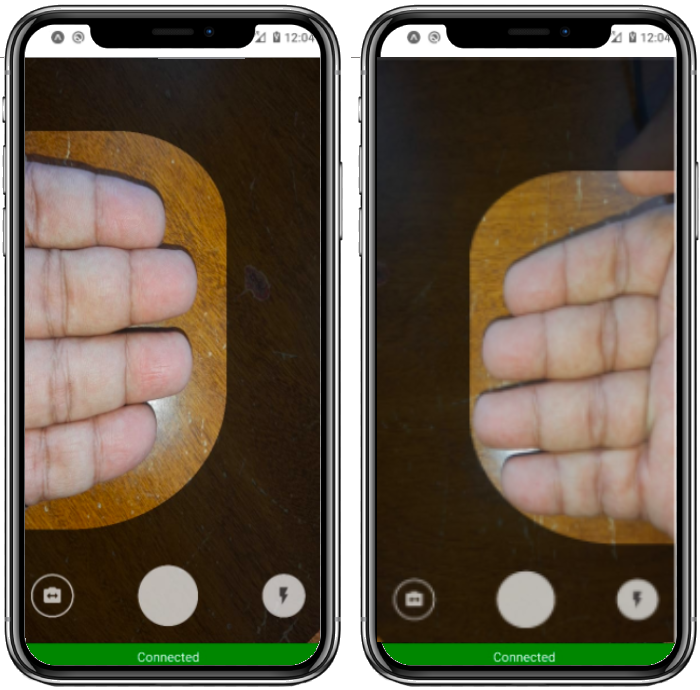}
\caption{Application used for acquiring four-finger images.}
\end{figure}

RidgeBase benchmark dataset has been collected over a period of 3 months from 88 participants. Contactless fingerprints were acquired using two smartphones, iPhone 11 and Google Pixel 5. We used an  application similar to Jawade et.al \cite{wifs_bhavin} for acquiring four-finger images using the two smartphones. As shown in figure 2, the application presents volunteers with a bounded region within which they can place their hand in an unconstrained manner. Corresponding contact-based fingerprints were acquired using \textit{Futronic FS64 EBTS} flat-bed fingerprint scanner. For each participant fingerprint images were collected over two sessions separated by atleast two weeks. There was significant gender, ethnicity ((East Asians, White Americans, African American, Filipino, Asian Indian and Filipino), race and age variation among subjects participating in the data collection.
This data collection was approved by the institutional IRB and the identities of the subjects involved in the data collection have been anonymized.

For each participant, contactless fingerprint images were acquired in three different lighting conditions and backgrounds namely (i) Indoor (ii) White Background and (iii) Outdoor. Each image was captured using flash and auto-focus. Contactless images were acquired using Apple iPhone 11 and Google Pixel 5 with resolutions (2016 x 4224) and (3024 x 4032) respectively. Across the 88 participants, we captured 280 contact-based four-finger images and 3374 contactless four-finger images. The dataset is further split using the distal segmentation approach as described in section 3.2. Table \ref{tab:stats} summarizes the dataset size and scope.

\subsection{Distal Segmentation Method}

Most fingerprint matching algorithms (such as Verifinger, \cite{polyudataset} \cite{ispfdv1}),  primarily work on distal fingerprints rather on the multi full-finger prints. To support compatibility with these algorithms and interoperability with existing datasets, we segment the four-finger images to extract distal phalanges. To produce pseudo bounding boxes for distal phalanges that can be then used to train an object detection model, we formulate a heuristic algorithm based on localization of convex defects.


We start by segmenting the background and the four-finger foreground. To perform this segmentation, we follow steps similar to \cite{wifs_bhavin}. First, we downsample the image and apply Gradcut algorithm using the guiding region presented to the user as a prior. We next apply morphological opening using kernels of size $(11,11)$ and $(5, 5)$ over the predicted grabcut mask $M$. Applying the up-scaled and Gaussian blurred mask over the original image gives us the segmented four-finger region. 

\begin{figure}
\label{imagesegmentation}
\includegraphics[scale=0.55]{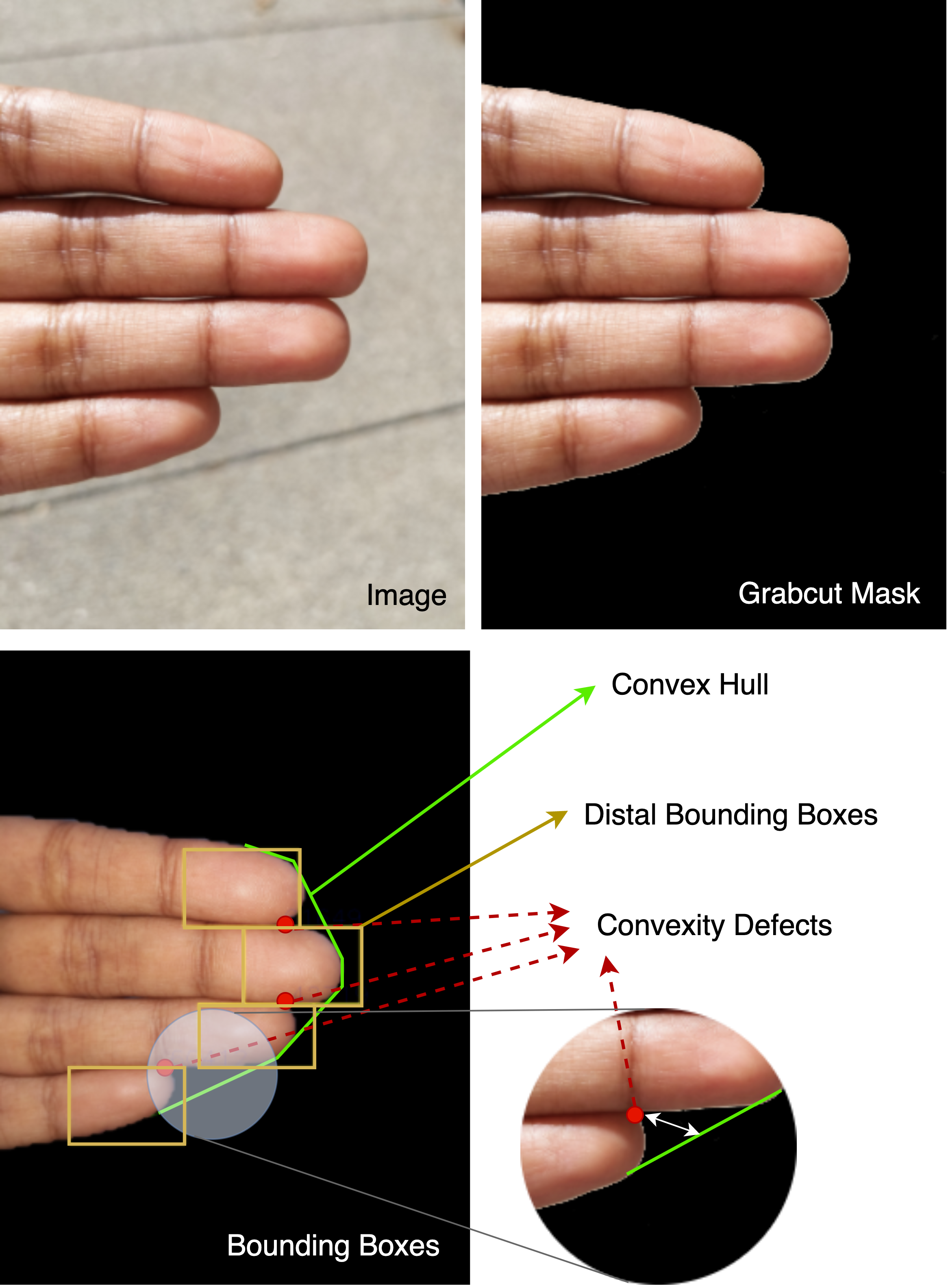}
\caption{Segmentation process for generating pseudo labels for finger images.}
\end{figure}

Next we find the convex hull $C$ for the segmented mask $M$ using  Sklansky's algorithm. Figure \ref{imagesegmentation} shows the convex hull over the four-finger region. For the set of images in the dataset that are acquired keeping the four fingers close to each other, the top-most point shared by any two fingers in contact must also be the farthest point on the perimeter of the segmented region from the convex hull. Under this premise, we detect top three farthest points (denoted by set $S$) from the convex hull (referred to as convexity defects). 

\[
S = \{(x_1, y_1),(x_2, y_2),(x_3, y_3)\}
\]

Next, we apply a set of empirically observed measures to generate bounding boxes for the four distals using the set $S$. We start by computing finger width using $y_2, y_3, y_4$.
\[
D_w = max((y3-y2),(y4-y3))
\]

Next, the following set of rules are used to predict bounding boxes around distals:
\begin{equation}
\begin{split}
& D_{TL_1} = (x_2 + 2*\alpha - \beta*D_w, y_2 - D_w) \\
& D_{BR_1} = (x_2 + 2*\alpha, y_2) \\
\end{split}
\end{equation}

\begin{equation}
\begin{split}
& D_{TL_2} = (x_3 + 4*\alpha - \beta*D_w , y_2) \\
& D_{BR_2} = (x_3 + 4*\alpha + 0.5*D_w, y_3) \\
\end{split}
\end{equation}

\begin{equation}
\begin{split}
& D_{TL_3} = (x_3 + 3*\alpha - D_w * \beta, y_3) \\
& D_{BR_3} = (x_3 + 3*\alpha, y_4) \\
\end{split}
\end{equation}

\begin{equation}
\begin{split}
& D_{TL_4} = (x_4 + 2*\alpha - D_w * \beta, y_4) \\
& D_{BR_4} = (x_4 + 2*\alpha, y_4 + D_w) \\
\end{split}
\end{equation}

Here, $\alpha = 1.5$ denotes an approximation of distal height to width ratio and $\beta$ is selected empirically as 50. 

We employ a visual selection method to pick 980 images that are perfectly annotated by the heuristic method, with 800 images being used for training and 180 images being used to test a FasterRCNN network for recognizing distal phalanges.
The $mAP @ IOU = 50\%$ of the trained FasterRCNN network is 95.7\%.
The final trained FasterRCNN is capable of detecting finger distals in images where the four fingers are not close to each other, despite the fact that it was trained with images with four fingers close to each other. 

\section{Tasks and Protocols}

We design the RidgeBase dataset to support three sets of tasks: (i) Single Finger Matching (or Distal-to-Distal matching) (ii) Four Finger Matching and (iii) Set based Distal Matching. Each task is further divided into contactless-to-contactless and contact-to-contactless verification and identification tasks. Unlike previous datasets, to ensure reproducibility, we provide fixed test evaluation pairs for each of the tasks. Below we provide detailed description of the tasks, evaluation protocols and associated train-test splits.

\subsection{Single Finger Matching}

Task 1 represents the single finger distal-to-distal matching scenario which is equivalent to traditional fingerprint matching. For task 1, we segment the distal phalanges for the whole dataset using the method described in section 3. So, for the 88 participants, the dataset consists of 704, (88 x 4 x 2) unique fingers. We select 200 unique fingers (classes) for the test set and 504 disjoint unique fingers (classes) for the training set. This protocol is for one-to-one fingerprint matching approaches, and considers each unique finger as a unique identity. This provides comparability and support for dataset augmentation with existing contactless fingerprint datasets that consist only of single finger images. In total, the test set consists of 2229 contactless finger distal images, and 200 contact-based fingerprints. The train set consists of 11255 contactless distal images and 916 contact-based fingerprints. For the contactless to contactless matching (CL2CL verification) task, we provide 24,83,106 test pairs for evaluation and for contact-to-contactless based matching (C2CL verification), we provide 4,54,716 test pairs.

\subsection{Four Finger Matching}

Task 2 represents the four-finger to four-finger matching scenario. Typically, multi-finger authentication is more robust than single finger authentication. This protocol promotes research in end-to-end trainable algorithms and feature fusion methods that can overcome distortion challenges of contactless images by utilizing identity features available in the entire four finger region. Here, we consider a hand (four-finger region) as a unique identity. For 88 participants, the task consists of 176 unique hands. We use 25 participants ($\sim$30\%) for test, and 63 participants for training (as in task 1). Therefore, task 2 consists of 50 unique four-finger images for test set and 126 unique four-finger images for train set. 

\subsection{Set-Based Matching}

To overcome the inconsistencies and distortions observed in real-time unconstrained capture of fingerprint images using smartphone camera, we introduce a set-based matching protocol. Set-based matching schemes have been previously used for face recognition \cite{januscvpr} where there is high intra-class variations. In task 3, each set consists of finger-distal images of the same finger under different backgrounds and lighting conditions and acquired using different devices in multiple sessions. For contactless-to-contactless distal matching, test split consists of 200 query sets and associated 200 gallery sets. On average each query-set consists of 4 samples, and each gallery-set consists of 5 samples. Similarly, for contact-to-contactless matching each gallery-set consists of 8 samples on average, and each query-set consists of 1 contact-based image. A robust feature fusion method developed to perform well on set-based matching protocol can greatly improve contactless matching performance in real-world where multiple images can be acquired from a continuous video.

\section{Quality Analysis of Fingerprints (NFIQ 2.0)}
\begin{figure}[t]
\includegraphics[scale=0.55]{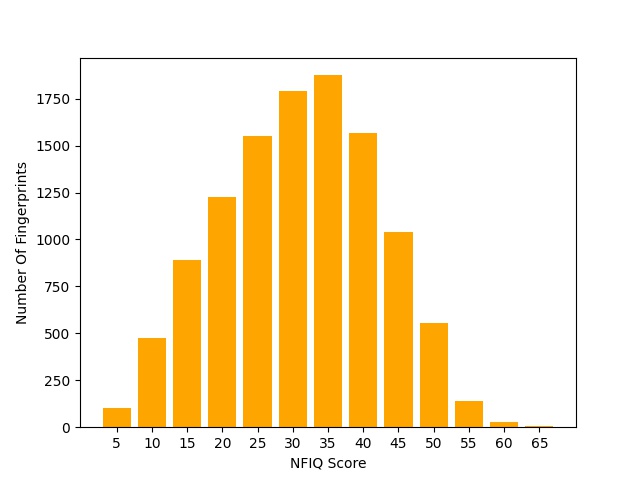}
\caption{NFIQ 2.0 score distribution of training split of RidgeBase dataset (Grayscaled)}
\label{fig:nfiq_8bit_500dpi_grayscaled_train}
\end{figure}

\begin{figure}
\includegraphics[scale=0.55]{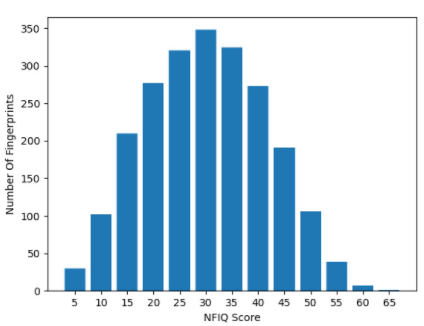}
\caption{NFIQ 2.0 score distribution of evaluation split of RidgeBase dataset (Grayscaled)}
\label{fig:nfiq_8bit_500dpi_grayscaled}
\end{figure}

\begin{figure}
\includegraphics[scale=0.55]{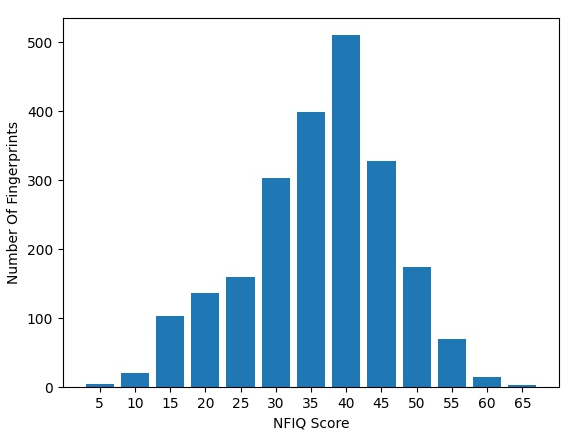}
\caption{NFIQ 2.0 score distribution for enhanced fingerprints (Enhanced \cite{anilj_enh}) in evaluation split}
\label{fig:nfiq_8bit_500dpi_anilenh}
\end{figure}

Figure \ref{fig:nfiq_8bit_500dpi_grayscaled} shows the distribution for fingerprint quality estimated using NFIQ 2.0 \cite{nfiq2} for the test-set split of Task 1 (only distals). All raw contactless distal images are gray scaled and converted to 8bit and 500 dpi before computing NFIQ scores. As can be observed from the distribution, a majority of fingerprints have NFIQ 2.0 scores in the range 20-45. Galbally et.al \cite{8987244} trained a bayes classifier for computing NFIQ 1.0 classes from NFIQ 2.0 values. Their learned mapping function \cite{8987244} can be summarized as:
\vspace{-1.0em}
\[
NFIQ 1 =
\begin{cases}
5 & \text{if $0 < NFIQ 2 \leq 5$} \\
3 & \text{if $6 < NFIQ 2 <= 35$} \\
2 & \text{if $36 < NFIQ 2 <= 45$} \\
1 & \text{if $46 < NFIQ 2 <= 100$} \\
\end{cases}
\]

where, NFIQ1 = 5 denotes worst quality images, and NFIQ1 = 1 denotes best quality images (NFIQ1=4 and NFIQ1=5 are treated as one unique class \cite{8987244}). Using this mapping function we observe that, for raw gray scale contactless images in RidgeBase test dataset 2.5\% images lie in NFIQ1 class 5, 76.7\% in class 3, 16.0\% in class 2 and 4.8\% in class 1.
Figure \ref{fig:nfiq_8bit_500dpi_grayscaled_train} shows the NFIQ2 score distribution for RidgeBase's training split. As it can be observed from Figure \ref{fig:nfiq_8bit_500dpi_grayscaled} and \ref{fig:nfiq_8bit_500dpi_grayscaled_train}, test set is representative of the training set in terms of raw fingerprint image quality distribution.
Figure \ref{fig:nfiq_8bit_500dpi_anilenh} shows NFIQ2 score distribution after enhancing contactless fingerprints using Hong. et.al's algorithm \cite{anilj_enh} to improve ridge clarity based on local ridge orientation and frequency.




\section{Experiments}
We evaluate baseline methods for verification (1:1) and Identification (1:N) tasks. We provide subject disjoint training and testing sets for all the tasks. Furthermore, the protocol also provides defined query and gallery templates for both verification and identification for Task 3.
\subsection{Metrics - Verification and Identification (1:N, 1:1), ROC and CMC}
  Methods are compared for verification task using EER (Equal Error rate), TAR(\%)@FAR=$10^{-2}$ (as in previous contactless matching works) and AUC. For Identification tasks, we report Rank($\%$)@1, Rank($\%$)@10, Rank($\%$)@50 and Rank($\%$)@100. Additionally, we compare methods using Receiver Operating Characterstic (ROC) and Cumulative Match Characterstic (CMC) for verification and identification respectively.

\subsection{Preprocessing and Enhancements}

We perform a set of pre-processing and ridge enhancement steps over the distal images segmented using the method described in section 3.2. We start by gray-scaling the contactless fingerprint images and then performing adaptive contrast enhancement with binary inversion. This is done to improve the ridge-valley contrast and account for the ridge inversion. We observe that due to variations in the focus over the distal region, directly enhancing the preprocessed contactless image leads to large number of spurious minutiae in the out of focus region. To address this, we perform adaptive Gaussian thresholding over the preprocessed image followed by a series of median blurs. This removes the out-of-focus regions of the distal image, leaving behind the sharp ridge pattern. Next, we enhance the fingerprint ridge pattern using the ridge frequency based enhancement method proposed by Hong et.al\cite{anilj_enh}.



\begin{table}[]
\label{CL2CL - 1:1 Verification}
\caption{CL2CL - 1:1 Verification}
\vspace{0.7em}
\begin{tabular}{lll}
\hline
\multicolumn{3}{c}{\textbf{Task 1 - Distal Matching}}                                                         \\ \hline
\multicolumn{1}{l|}{Metric}                               & \multicolumn{1}{c|}{Verifinger} & AdaCos (CNN) \\ \hline
\multicolumn{1}{l|}{EER (\%)}                       & \multicolumn{1}{c|}{19.7}           &  \multicolumn{1}{c}{21.3}            \\
\multicolumn{1}{l|}{TAR(\%)@FAR=$10^{-2}$} & \multicolumn{1}{c|}{63.3}           &   \multicolumn{1}{c}{61.2}           \\
\multicolumn{1}{l|}{AUC(\%)} & \multicolumn{1}{c|}{89.3}           &      \multicolumn{1}{c}{87.7}        \\ 
\hline
\multicolumn{3}{c}{\textbf{Task 2 - Four Finger Matching}}                                                    \\ \hline
\multicolumn{1}{l|}{Metric}                               & \multicolumn{1}{l|}{Verifinger} & AdaCos (CNN) \\ \hline
\multicolumn{1}{l|}{EER (\%)}                       & \multicolumn{1}{c|}{13.1}           &     \multicolumn{1}{c}{14.8}         \\
\multicolumn{1}{l|}{TAR@FAR=$10^{-2}$} &  \multicolumn{1}{c|}{79.8}           &             
\multicolumn{1}{c}{70.9}
\\ 
\multicolumn{1}{l|}{AUC(\%)} & \multicolumn{1}{c|}{92.1}           &   \multicolumn{1}{c}{92.6}           \\ 
\hline
\multicolumn{3}{c}{\textbf{Task 3 - Set based Distal Matching}}                                                      \\ \hline
\multicolumn{1}{l|}{Metric}                               & \multicolumn{1}{l|}{Verifinger} & AdaCos (CNN) \\ \hline
\multicolumn{1}{l|}{EER (\%)}                       & \multicolumn{1}{c|}{7.90}           &     \multicolumn{1}{c}{9.5}         \\
\multicolumn{1}{l|}{TAR(\%)@FAR=$10^{-2}$} & \multicolumn{1}{c|}{86.1}           &  \multicolumn{1}{c}{86.5}              \\ 
\multicolumn{1}{l|}{AUC(\%)} & \multicolumn{1}{c|}{95.3}           &   \multicolumn{1}{c}{96.3}           \\ 
\hline
\end{tabular}
\end{table}

\begin{table}[]
\label{CL2CL - 1:N Identification}
\caption{CL2CL - 1:N Identification}
\vspace{0.7em}
\begin{tabular}{lllll}
\hline
\multicolumn{5}{c}{\textbf{Task 1 - Distal Matching}}                                                                                 \\ \hline
\multicolumn{1}{l|}{Method}             & \multicolumn{1}{l|}{R@1} & \multicolumn{1}{l|}{R@10} & \multicolumn{1}{l|}{R@50} & R@100 \\ \hline
\multicolumn{1}{l|}{Verifinger}   & \multicolumn{1}{l|}{85.2}    & \multicolumn{1}{l|}{91.4}     & \multicolumn{1}{l|}{93.8}     &  95.4     \\
\multicolumn{1}{l|}{AdaCos (CNN)} & \multicolumn{1}{l|}{81.9}    & \multicolumn{1}{l|}{89.5}     & \multicolumn{1}{l|}{94.1}     &  95.9     \\ \hline
\multicolumn{5}{c}{\textbf{Task 2 - Four Finger Matching}}                                                                            \\ \hline
\multicolumn{1}{l|}{Method}             & \multicolumn{1}{l|}{R@1} & \multicolumn{1}{l|}{R@10} & \multicolumn{1}{l|}{R@50} & R@100 \\ \hline
\multicolumn{1}{l|}{Verifinger}   & \multicolumn{1}{l|}{94.1}    & \multicolumn{1}{l|}{99.0}     & \multicolumn{1}{l|}{99.8}     & 100.0      \\
\multicolumn{1}{l|}{AdaCos (CNN)} & \multicolumn{1}{l|}{91.5}    & \multicolumn{1}{l|}{97.3}     & \multicolumn{1}{l|}{99.6}     &   99.8    \\ \hline
\multicolumn{5}{c}{\textbf{Task 3 - Set Based Distal Matching}}                                                                              \\ \hline
\multicolumn{1}{l|}{Method}             & \multicolumn{1}{l|}{R@1} & \multicolumn{1}{l|}{R@10} & \multicolumn{1}{l|}{R@50} & R@100 \\ \hline
\multicolumn{1}{l|}{Verifinger}   & \multicolumn{1}{l|}{91.5}    & \multicolumn{1}{l|}{99.5}     & \multicolumn{1}{l|}{100.0}     & 100.0      \\
\multicolumn{1}{l|}{AdaCos (CNN)} & \multicolumn{1}{l|}{86.5}    & \multicolumn{1}{l|}{99.0}     & \multicolumn{1}{l|}{100.0}     &  \multicolumn{1}{l}{100.0}     \\ \hline
\end{tabular}
\vspace{-1em}
\end{table}

\subsection{Baselines}
We present evaluations on the RidgeBase dataset using the commercial-off-the-shelf (COTS) Verifinger matcher and the CNN based deep metric learning method proposed in \cite{wifs_bhavin}. 
To generate ISO templates using Verifinger 12.0 we first preprocess and enhance fingerprints using the algorithm described in section 6.2, and then convert the fingerprints to 8 bit 500dpi images. 

For the second baseline, we evaluate the AdaCos based branch as described in \cite{wifs_bhavin}. The model takes a channel sequenced, enhanced and grayscaled image as input followed by Densenet 161 representation extractor optimized with adaptive scaling cosine (AdaCos) loss \cite{adacos}. We first pretrain the network using 50,000 synthetic fingerprints generated using the Anguli \footnote{Anguli: https://dsl.cds.iisc.ac.in/projects/Anguli/index.html} Fingerprint Generator and then fine tune over RidgeBase. We use 2000 images out of the 11,252 images for validation and the remaining 9,252 images for training. Results reported for both baselines are over the RidgeBase test split. 
For task 2 and task 3, we segment the distal phalanges and perform score fusion using sum rule. End-to-End training of four-finger region and association-based feature pooling are left for future exploration.

\begin{figure}
\includegraphics[scale=0.7]{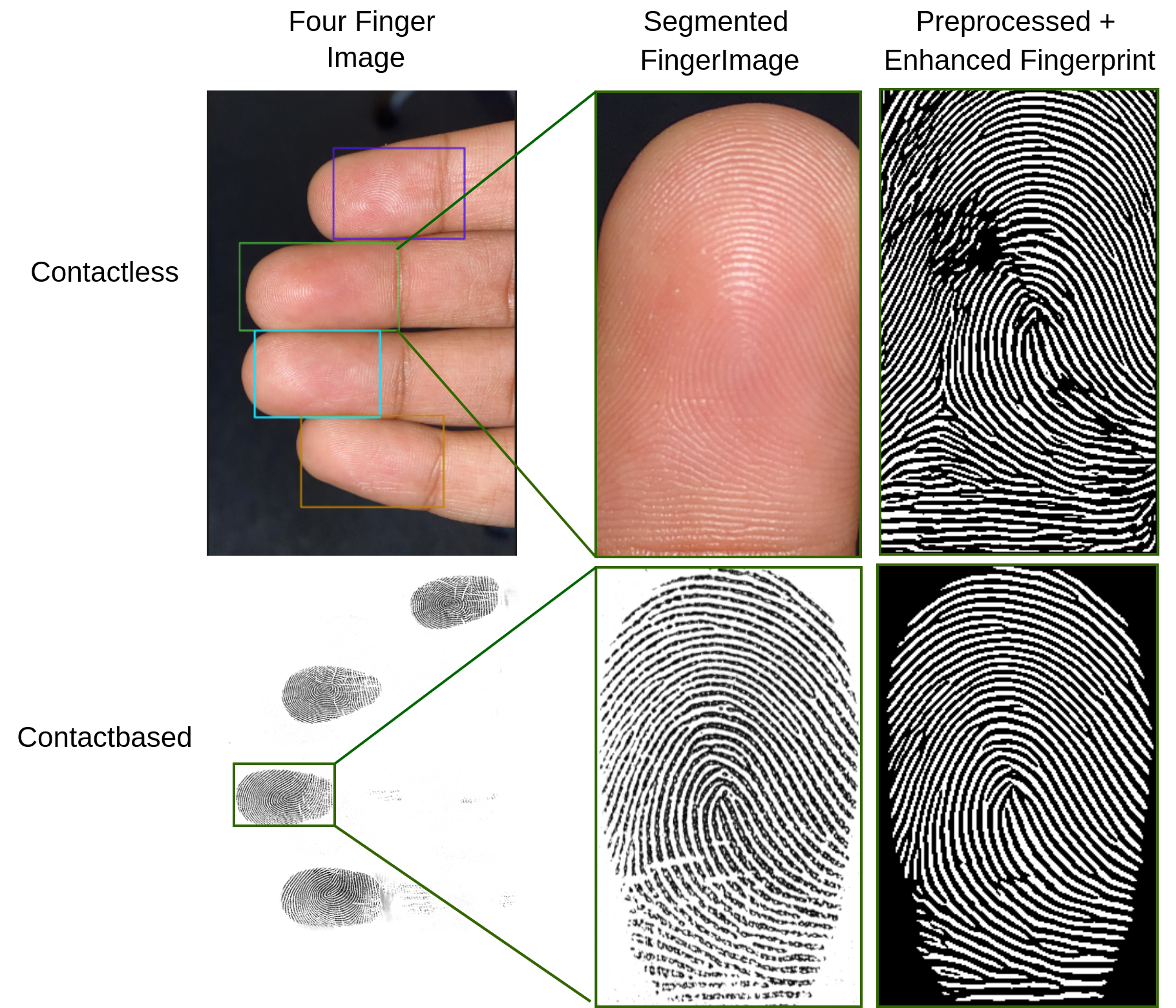}
\caption{Contactless and Contact Four-Finger, Segmented, and Enhanced Fingerprint samples}
\label{fig:dataset_sample_images}
\end{figure}



\begin{figure}[t]
\label{fig:cl2cl_roc}
\includegraphics[scale=0.27]{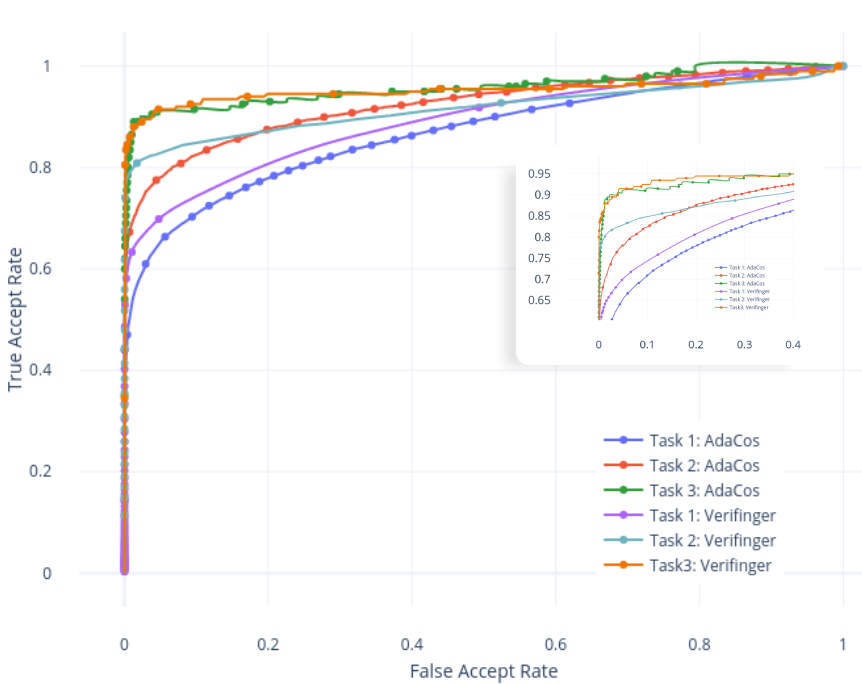}
\caption{1:1 Verification ROCs for Contactless-2-Contactless}
\end{figure}

\begin{figure}[t]
\label{fig:cl2cl_cmc}
\includegraphics[scale=0.37]{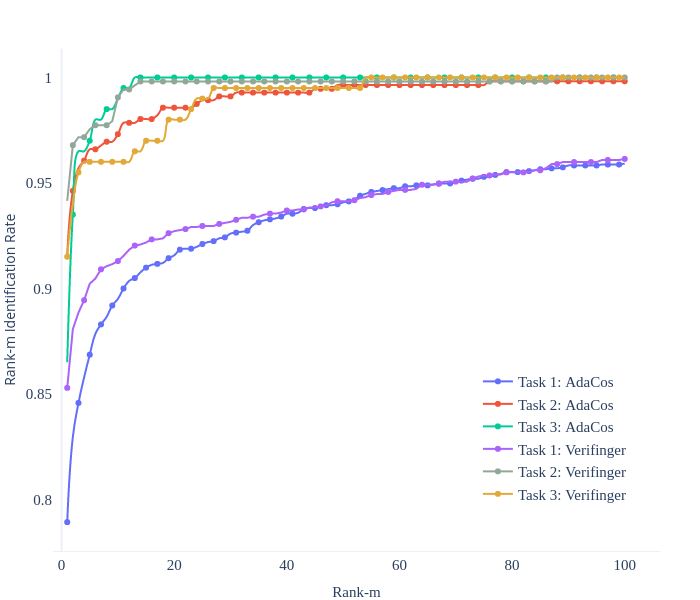}
\caption{1:N Identification CMCs for Contactless-2-Contactless}
\end{figure}

\begin{figure}[t]
\includegraphics[scale=0.3]{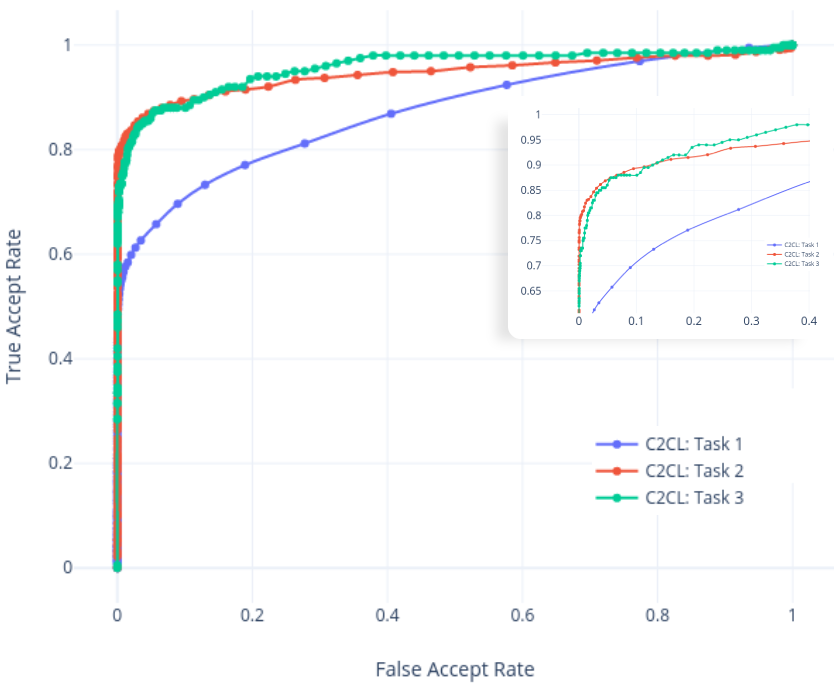}
\caption{1:1 Verification ROCs for Contact-2-Contactless (Method: Verifinger)}
\label{fig:verifinger_roc_c2cl}
\end{figure}

\begin{figure}[t]
\includegraphics[scale=0.41]{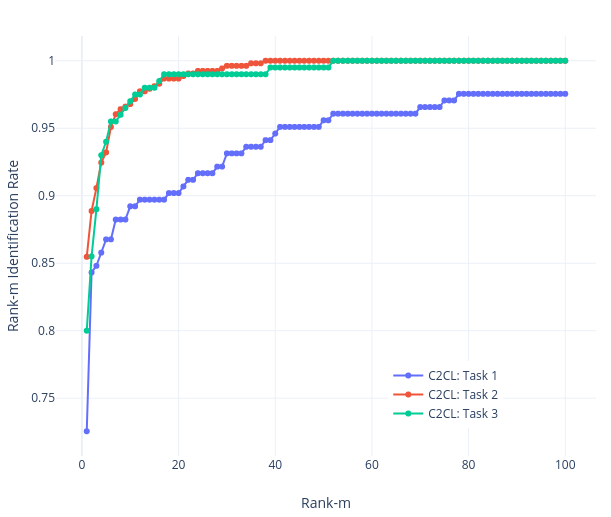}
\caption{1:N Identification CMCs for Contact-2-Contactless (Method: Verifinger)}
\label{fig:verifinger_cmc_c2cl}
\end{figure}

\begin{table}
\centering
\caption{C2CL - 1:1 Verification - Verifinger}
\vspace{0.5em}
\label{C2CL - 1:1 Verification}
\begin{tabular}{l|c|c|c} 
\hline
             \multicolumn{1}{l|}{\textbf{Task}}             & \multicolumn{1}{l|}{EER (\%)} & \multicolumn{1}{l|}{TAR(\%)@FAR=${10^{-2}}$} & \multicolumn{1}{l}{AUC(\%)}  \\ 
\hline
\textbf{\textbf{Task 1}}  & 18.9                          & 57.6                                          & 87.2                         \\
\textbf{\textbf{Task 2~}} & 11.3                          & 82.4                                          & 95.1                         \\
\textbf{\textbf{Task 3~}} & 10.7                          & 79.3                                          & 95.7                         \\
\hline
\end{tabular}
\end{table}

\begin{table}
\centering
\caption{C2CL - 1:N Identification - Verifinger}
\vspace{0.5em}
\label{C2CL - 1:N Identification}
\begin{tabular}{l|l|l|l|l} 
\hline
\textbf{Task}               & R@1                  & R@10                 & R@50                 & R@100  \\ 
\hline
\textbf{Task 1}               & 72.5                 & 89.2                 & 95.5                 & 97.5   \\
\textbf{Task 2}               & 85.4                 & 96.7                 & 100.0                  & 100.0    \\
\textbf{Task 3}               & 80.0                 & 97.0                & 99.5.                 & 100.0   \\ 
\hline
\end{tabular}
\vspace{-2em}
\end{table}

\subsection{Results}

Table \ref{CL2CL - 1:1 Verification} and \ref{C2CL - 1:1 Verification} report the verification results for contactless-to-contactless matching and contact-to-contactless matching respectively for all three tasks i.e. Distal Matching, Four Finger Matching, and Set Based Distal Matching. Figure \ref{fig:cl2cl_roc} and \ref{fig:verifinger_roc_c2cl} shows the receiver operating characteristic (ROC) for all three tasks. Table \ref{CL2CL - 1:N Identification} and \ref{C2CL - 1:N Identification} report the Identification results for the contactless-to-contactless matching and contact-to-contactless matching task respectively. Figure \ref{fig:cl2cl_cmc} and \ref{fig:verifinger_cmc_c2cl} show the Cumulative Match Curve for the identification rate. Based on the performance evaluation of both the widely used COTS verifinger and the CNN based method, we observe that RidgeBase is more challenging than other existing contactless fingerprint datasets, and hence motivates further innovation in contactless fingerprint matching algorithms.



\section{Conclusion}

In this work, we have proposed a novel smartphone based contactless fingerprint matching dataset. RidgeBase, a multi-use full-finger dataset, will help advance new avenues for contactless fingerprint matching, promoting methods that could leverage different parts from the four-finger region for matching. With the set-based matching protocol introduced along with RidgeBase, novel contactless fusion algorithms can be investigated to achieve better query-set to gallery-set matching performance. Along with this dataset, we release the cross-platform app developed to collect the fingerphotos.

\section{Acknowledgement}
This work was conducted at the Center for Unified Bio-
metrics and Sensors (CUBS) at the University at Buffalo and
was supported by the Center for Identification Technology Research (CITeR) and the National Science Foundation through
grant \#1822190.


{\small
\bibliographystyle{ieee}
\bibliography{egbib}
}

\end{document}